\documentclass[letterpaper, 10 pt, conference]{ieeeconf}  

\IEEEoverridecommandlockouts                              

\usepackage{cite}

\usepackage{bbding}
\usepackage{amsmath,amssymb,amsfonts}
\usepackage{algorithmicx}
\usepackage{algorithm}
\usepackage{graphicx}
\usepackage{textcomp}
\usepackage{xcolor}
\usepackage{booktabs}
\usepackage{comment}
\usepackage{pifont}

\usepackage{subfigure}
\usepackage{comment,stmaryrd}
\usepackage{algpseudocode}

\makeatletter
\let\NAT@parse\undefined
\makeatother
\usepackage{hyperref}
\hypersetup{
	colorlinks=true,
	linkcolor=blue,
	filecolor=magenta,      
	urlcolor=cyan,
	citecolor=blue,
	bookmarks=true,
}

\title{\LARGE \bf
Optimal Smoothing Distribution Exploration for Backdoor Neutralization in Deep Learning-based Traffic Systems}

\author{Yue Wang$^{\dagger,1}$, Wenqing Li$^{2}$, Michail Maniatakos$^{3}$ and Saif Eddin Jabari$^{4}$
\thanks{$^{1}$Yue Wang is a Ph.D. Student in the Department of Electrical Engineering,
        Tandon school of Engineering, New York University, Brooklyn, NY 11201, USA
        \url{yw3576@nyu.edu}}%
\thanks{$^{2}$Wenqing Li is a Post-doc associate in the Division of Engineering, New York University Abu Dhabi, Saadiyat Island 129188, UAE
        \url{wl54@nyu.edu}}%
\thanks{$^{3}$Michail Maniatakos is a faculty member in the Division of Engineering, New York University Abu Dhabi, Saadiyat Island 129188, UAE
        \url{michail.maniatakos@nyu.edu}}%
\thanks{$^{4}$Saif Eddin Jabari is a faculty member in the Division of Engineering, New York University Abu Dhabi, Saadiyat Island 129188, UAE
        \url{sej7@nyu.edu}}%
   \thanks{$\dagger$: Corresponding author.}
}

\begin{document}

\maketitle
\thispagestyle{empty}
\pagestyle{empty}


\begin{abstract}
 Deep Reinforcement Learning (DRL) enhances the efficiency of Autonomous Vehicles (AV), but also makes them susceptible to backdoor attacks that can result in traffic congestion or collisions. Backdoor functionality is typically incorporated by contaminating training datasets with covert malicious data to maintain high precision on genuine inputs while inducing the desired (malicious) outputs for specific inputs chosen by adversaries. Current defenses against backdoors mainly focus on image classification using image-based features, which cannot be readily transferred to the regression task of DRL-based AV controllers since the inputs are continuous sensor data, i.e., the combinations of velocity and distance of AV and its surrounding vehicles. Our proposed method adds well-designed noise to the input to neutralize backdoors. The approach involves learning an optimal smoothing (noise) distribution to preserve the normal functionality of genuine inputs while neutralizing backdoors. By doing so, the resulting model is expected to be more resilient against backdoor attacks while maintaining high accuracy on genuine inputs. The effectiveness of the proposed method is verified on a simulated traffic system based on a microscopic traffic simulator, where experimental results showcase that the smoothed traffic controller can neutralize all trigger samples and maintain the performance of relieving traffic congestion. 
\end{abstract}

\section{Introduction}

The emergence of autonomous vehicles (AVs) has the potential to revolutionize transportation systems by increasing mobility and safety. The development of AVs can help solve long-standing transportation challenges, including traffic congestion \cite{DBLP:journals/corr/SternCMBBCHHPWP17,vahidi2018energy}. To approximate the highly non-linear nature of driving, Deep Neural Networks (DNNs) have been widely adopted in AVs. However, recent research has shown that DNNs are vulnerable to backdoor attacks \cite{DBLP:journals/access/GuLDG19,DBLP:conf/iclr/NguyenT21,DBLP:conf/uss/BagdasaryanS21}, where an adversary can manipulate the DNN to embed backdoor functionalities that can cause misclassifications for specific adversary-chosen inputs. The backdoor functionality is usually implanted by poisoning training datasets with malicious inputs having stealthy trigger patterns and incorrect target labels. Recent work demonstrated that DRL-based traffic controllers of AVs are also susceptible to stealthy backdoor attacks \cite{DBLP:journals/tifs/WangSLMJ21}, where an adversary could misclassify controller outputs of a backdoored AV by adding malicious triggers based on traffic physics with the sensor inputs. They proposed two types of attacks: (1) congestion attack - causing traffic congestion, and (2) insurance attack - causing the AV to crash into the vehicle in front. Therefore, it is important to develop robust defenses against backdoor attacks to ensure the safety and reliability of AVs.

The current defense methods \cite{DBLP:conf/ccs/LiuLTMAZ19,DBLP:conf/sp/WangYSLVZZ19,DBLP:journals/dt/SarkarAM20,DBLP:conf/sp/ChouTP20}  against backdoors in DNNs concentrate primarily on image classification tasks and can be divided into to categories. Techniques pertaining to one category (such as Neural Cleanse \cite{DBLP:conf/sp/WangYSLVZZ19}, Sentinet \cite{DBLP:conf/sp/ChouTP20} and ABS \cite{DBLP:conf/ccs/LiuLTMAZ19}) defense against backdoors by detecting specific trigger patterns that significantly impact the model classification performance. However, they might be breakable when facing advanced attacks \cite{wong2018provable,raghunathan2018certified}), and they cannot be transferred to DRL-based AV controllers  (as these controllers use continuous sensor data as inputs and the output are continuous values instead of class labels). Alternatively, randomized smoothing (RS) \cite{cohen2019certified} constructs a smoothed model via adding noise to the inputs and outputting the expected perturbed inputs. This increases the robustness of the model with potentially embedded backdoors. RS does not depend on a specific form of triggers and is effective whenever the triggers satisfy some general conditions. However, it impacts the accuracy of the model, and it is still limited to classification problems rather than regression problems.

Defense methods against backdoor attacks in sensor-based traffic control systems, which use streaming sensor data such as velocities and positions as input for the model, have not yet been fully explored. Therefore, we propose an approach that draws inspiration from randomized smoothing (RS) to learn an optimal smoothing (noise) distribution that balances model accuracy and robustness. Specifically, we learn an unnormalized density function of noise parameters that connects the noise to a well-defined metric that reveals the performance of the smoothed model. We can subsequently generate the desired noise by sampling from this density function. By doing so, we can maintain the model functionality while rendering any backdoor present in the model ineffective.
 The contributions of this work are: 
\begin{enumerate}
    \item We systematically explore the optimal smoothing (noise) distribution for regression tasks of DRL-based AV controllers based on sensor-values;
    \item We learn an unnormalized density function and propose a sampling strategy to generate desired noise, where we ensure that the optimal (noise) sample with maximum probability can be generated;
    \item  Experimental evaluation on AV controllers targeted for injecting backdoors demonstrates that the proposed method outperforms existing state-of-the-art defense methods.
\end{enumerate}


The rest of this paper is organized as follows: Section \ref{pre} presents the basic notions of backdoor attacks on AV controllers and randomized smoothing. 
In Section \ref{explore_noise}, we provide details of our methodology for learning the optimal smoothing (noise) distributions for backdoor neutralization. 
In Section \ref{s:Experimental}, we provide experiments and conclude the paper in 
Section \ref{s:conclusion}.

\section{Preliminaries}\label{pre}

In this section, we will briefly revisit the general notions of backdoor attacks (on AV controllers) and randomized smoothing. 

\subsection{Backdoor attacks on AV controllers}
The AV controllers targeted in this work are based on DRL and are primarily used to relieve traffic congestion. They take the traffic state as input, e.g., velocities and positions, and output the command actions for the AV, e.g., speed changes. Backdoors on AV controllers aim to deliberately compromise the controller and produce a malicious behavior, which generates attacker-designed actions for certain trigger inputs \cite{wang2020stopandgo}. The trigger inputs and desired actions are sampled from a distribution that adheres to the principles of traffic physics.

\subsection{Randomized smoothing}
Randomized smoothing is an effective way to achieve robustness by constructing a smoothed model. Consider a controller function  $\mathcal{M}:\mathbb{R}^d\rightarrow\mathbb{R}$, randomized smoothing builds a smoothed controller $\mathcal{F}$ by perturbing the inputs with  Gaussian noise, and outputting the expected results over the perturbed inputs. The smoothed controller is constructed as follows:
\begin{equation}
    \mathcal{F}(x) = \mathbb{E}_{\epsilon \sim \mathcal{N}(0,\Sigma)}[\mathcal{M}(\mathbf{x+\epsilon})]
\end{equation}
where $\mathcal{N}(0, \Sigma)$ represents a Gaussian distribution, and $\Sigma$ is a diagonal matrix and each diagonal element $\sigma_i^2$ is the variance of each element in the inputs. In this sense, the output of perturbed input can maintain unchanged under mild conditions,
\begin{equation}
    \mathcal{F}(x+\varepsilon)=\mathcal{F}(x)
\end{equation}
with the magnitude of perturbation $\varepsilon$ smaller than some threshold.

It should be noted that there exists a trade-off between the model accuracy (i.e., accuracy with respect to genuine inputs) and model robustness (i.e., robustness with respect to perturbed inputs). An increase in the variance of Gaussian noise, for instance, will lead to a decrease in accuracy.

\section{Methodology}\label{explore_noise}
 The objective of the proposed approach is to acquire the optimal Gaussian noise, or the optimal parameters of Gaussian noise, that strikes a balance between the accuracy of genuine inputs and the resilience against perturbed inputs. To accomplish this, we learn a value function of noise parameters that establishes a link between the parameters of Gaussian noise and a clearly defined metric that reflects the performance of the smoothed model or controller. This value function is essentially an unnormalized probability density function (pdf) in the sense that the values of metric are unnormalized probabilities. Subsequently, desired noise parameters can be generated according to this density function. In this case, the backdoor in AV controllers can be rendered ineffective when exposed to trigger samples, and the system performance can be preserved in the normal (non-attacked) setting.
\subsection{Stability to trigger sensitivity ratio}

We measure the performance of the smoothed controller based on the above two expectations to define the best parameters of Gaussian noise. 
The noise parameters $\Sigma$ contain the variances of Gaussian noise. The inputs should be smoothed with the corresponding noise parameters such that the functionality of the controller is not compromised. This means that the controller should maintain a high average velocity and stabilize the system when in a clean environment. The system \emph{stability} metric $r_1$ is defined based on the mean $v_{avg}$ and the standard deviation $v_{std}$ of velocities of all vehicles in the traffic system,
\begin{equation}
    r_1 = v_{avg}/v_{std}
\end{equation}
Large $r_1$ show that the traffic system runs in a stable state with high average velocity.

The second requirement for the system is that it should be insensitive to triggers. This implies that after the smoothing process, trigger samples should have comparable characteristics to clean samples, resulting in the smoothed model producing similar outputs for both trigger and clean samples. Prior research \cite{wang2022pidan} has demonstrated that, for classification problems, representations from the hidden layer of neural networks are desirable to reside in the same linear latent subspace for samples that share similar characteristics. This principle also applies to regression problems when analyzing representations in high-dimensional kernel spaces, wherein the projections of representations for samples that share similar characteristics (i.e., from the same subspace) exhibit similar scales. To assess subspace affiliation, a projection-based feature $\mathbf{w}$ can be calculated, whereby features corresponding to samples from the same subspace are uni-modal and features  associated with samples from distinct subspaces are at least bi-modal. Subsequently, the likelihood ratio test can be leveraged to assess whether the feature is uni-modal or not. With respect to this metric, we generate several trigger samples from the trigger distributions learned using the same way as \cite{wang2020stopandgo}, even though obtaining the exact trigger samples employed to poison the controller is infeasible. This metric of \emph{trigger sensitivity} of the system is defined as follows: 

\begin{equation}
    r_2 = \frac{|J-J_{\mathrm{clean}}|} {|J_{\mathrm{clean}}|}
\end{equation}
where $J = -2\mathrm{log}\frac{\mathcal{L}(\mathbf{w}|\mathbf{H}_0)}{\mathcal{L}(\mathbf{w}|\mathbf{H}_1)}$ is a (log) likelihood ratio,  $\mathcal{L}(\mathbf{w}|\mathbf{H}_0)$ denotes the estimated likelihood of $\mathbf{w}$ under the null hypothesis that $\mathbf{w}$ is drawn from a uni-modal distribution, and $\mathcal{L}(\mathbf{w}|\mathbf{H}_1)$ represents the likelihood of $\mathbf{w}$ under the alternative hypothesis that $\mathbf{w}$ is drawn from a bi-modal distribution. $J$ is calculated from the mixture of clean and trigger representations. For $J_{\mathrm{clean}}$, the corresponding projection-based feature $\mathbf{w}$ is calculated from clean-only representations.

The ultimate measure of evaluation is expressed as the \emph{stability to trigger sensitivity ratio}, denoted as $r = \frac{r_1}{r_2}$. A higher value for this ratio suggests that the parameters used are more optimal.

\subsection{Value function learning and optimal parameter generation}
We denote by $x$ the noise parameter, and let $\tilde{p}_{\theta}(x)$ be the value function that connects noise parameters with ratio of stability to trigger sensitivity, where $\theta$ denotes the parameter of value function. The value function $\tilde{p}_{\theta}(x)$ can be regarded as an unnormalized probability density function with respect to $x$ in that the value of ratio denotes the unnormalized probability of $x$. The normalized density function, termed $p_{\theta}(x)$, is given by
\begin{equation}
    p_{\theta}(x)=\frac{\tilde{p}_{\theta}(x)}{Z(\theta)}
\end{equation}
where $Z(\theta)=\int \tilde{p}_{\theta}(x) \mathrm{d} x$ is the partition function.  Since the analytic form of $p_{\theta}(x)$ is often unavailable and integrals including $Z(\theta)$ are typically intractable \cite{hyvarinen2005estimation,martino2017metropolis}, we learn the value function (a.k.a., unnormalized density function) $\tilde{p}_{\theta}$, and generate optimal noise parameters by sampling from this unnormalized density function.  

We leverage a neural network to represent the value function $\tilde{p}_{\theta}$, where the input to the neural network is noise parameters (i.e., $x$) from a uniform distribution. The output is the corresponding ratio value, i.e., $\tilde{p}_{\theta}(x)$, and hence $\theta$ is the set of weights of the neural network. The loss function of the neural network is given as
\begin{equation}
    \mathcal{L}:=\mathbb{E}[\|\tilde{p}_{\theta}(x)-r(x)\|^2]+\lambda \|\theta\|^2
    \label{loss}
\end{equation} 
where $r(x)$ is the metric value associated with the input (a.k.a. parameter $x$), which is obtained by interacting with the traffic system, i.e., applying the smoothed controller based on $x$ to the traffic system. The second term with regularization parameter $\lambda$ aims to avoid overfitting. 

Subsequently, the optimal  parameter can be generated by sampling from this unnormalized density function. The basic idea of the sampling scheme goes as follows: one can generate a sample value from $\tilde{p}_{\theta}(x)$ by instead sampling from another distribution $q(x)$ and accept them based on some rule, and repeating the draws from $q(x)$ until a value is accepted.

The specific sampling strategy is detailed in what follows.
 \begin{enumerate}
     \item Given the value function (unnormalized density function) $\tilde{p}_{\theta}(x)$, choose a proposal distribution $q(x)$, i.e., a uniform distribution $\mathcal{U}(0,1)$.
     \item Generate a sample from a uniform distribution, i.e., $t\sim\mathcal{U}(a,1)$, where $a$ is a predefined  constant satisfying $a\in[0,1]$.
     \item Generate a sample from the proposal distribution $q(x)$, i.e., $x\sim q(x)$, and calculate the ratio
     \begin{equation}      \alpha=\frac{\tilde{p}_{\theta}(x)-aMq(x)}{Mq(x)(1-a)},
     \end{equation}
     where $M$ is a constant, finite bound on $\tilde{p}_{\theta}(x)/q(x)$, i.e., $$M=\underset{x}{\mathrm{sup}}\ \tilde{p}_{\theta}(x)/q(x).$$
    
     \item Check whether $t\leq \alpha$ or not: accept $x$ as a sample drawn from $\tilde{p}_{\theta}(x)$ if it holds, and then go to next step; reject $x$ if not and return to step 2). 
     \item Update constant $a$ as $$a=\mathrm{min}\{\tilde{p}_{\theta}(x)/Mq(x),1\}$$ with $x$ representing the accepted sample in step 4), and return to step 2).
     \item Accept $x$ as the optimal sample (a.k.a., noise parameter) if no more sample could be generated from $\mathcal{U}(a,1)$.
 \end{enumerate}
\begin{figure}[t]
	\centering
		\centering
		\includegraphics[scale=1]{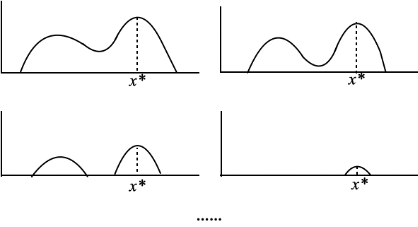}
	\caption{Illustration of the the unnormalized distribution $\tilde{p}_{\theta}(x)$ with increasing parameter $a$ ($a$ increases from top left to bottom right). We observe that although the distribution changes as $a$ changes, the point with maximum probability keeps unchanged.}
	\label{sp}
\vspace{-0.2in}
\end{figure}
\begin{figure}[t]
	\centering
		\centering
		\includegraphics[scale=1]{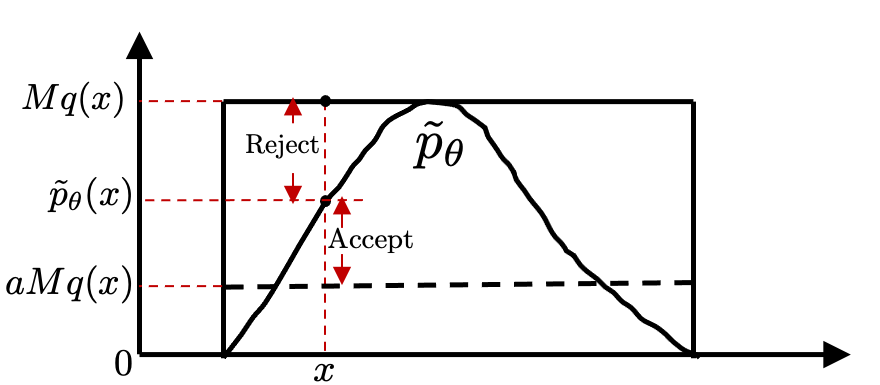}
	\caption{Illustration of rejecting or accepting $x$ as a sample drawn from $\tilde{p}_{\theta}$ with the extra condition $\tilde{p}_{\theta}(x)\geq aMq(x)=a\ \mathrm{sup}_x\ \tilde{p}_{\theta}(x)$.}
	\label{illup}
\vspace{-0.2in}
\end{figure}
 In this way, we demonstrate that the probability of $x$ provided it has been accepted in each iteration, is positive related to the normalized probability $p_{\theta}(x)$. This implies the sample (noise parameter) with the maximum probability of $p_{\theta}(x)$ (optimal performance) can be generated, which is illustrated in Fig.~\ref{sp}.
 
 \emph{More specifically, let us denote $\Omega$ the event that a sample $x$, drawn from $q(x)$ has been accepted in each iteration with respect to $a$. We denote by $\pi(x|\Omega)$ the probability of $x$, provided it has been accepted. We have $\pi(x|\Omega)=\beta p_{\theta}(x)+\gamma$, where $\beta>0$ and $\gamma$ are constant, and $p_{\theta}(x)$ is the normalized density function.}

 \textbf{Analysis:} According to Bayes’ theorem we have 
 \begin{equation}
     \pi(x|\Omega)=\frac{\pi(\Omega|x)\pi(x)}{\pi(\Omega)}
 \end{equation}
where $\pi(x)=q(x)$ is the distribution where sample drawn from, i.e., proposal distribution, and $\pi(\Omega|x)$ is the probability of accepted provided $x$, which is given by 
\begin{equation}
    \pi(\Omega|x)=\alpha=\frac{\tilde{p}_{\theta}(x)-aMq(x)}{Mq(x)(1-a)}
\end{equation}
Then, the probability of event $\Omega$ is given by 
 \begin{multline}
\pi(\Omega)=\int \pi(x,\Omega)dx =\int \pi(\Omega|x)q(x)dx\\=\int\frac{\tilde{p}_{\theta}(x)-aMq(x)}{Mq(x)(1-a)}q(x)dx=\frac{1}{M(1-a)}(Z(\theta)-aM)
 \end{multline}
 The last equality follows from $\int\tilde{p}_{\theta}(x)dx=Z(\theta)$ and $\int q(x)dx=1$.
 
This leads to
\begin{equation}  \pi(x|\Omega)=\frac{\pi(\Omega|x)q(x)}{\pi(\Omega)}=\frac{Z(\theta)}{Z(\theta)-aM}p_{\theta}(x)-\frac{aM}{Z(\theta)-aM}q(x)
\end{equation}
Since $q(x)=\frac{1}{1-a}$ (a uniform distribution in term of $\mathcal{U}(a,1)$), we have
\begin{equation}
    \pi(x|\Omega)=\frac{Z(\theta)}{Z(\theta)-aM}p_{\theta}(x)-\frac{aM(1-a)}{Z(\theta)-aM}
\end{equation}
Let $\beta=\frac{Z(\theta)}{Z(\theta)-aM}$ and $\gamma=-\frac{aM(1-a)}{Z(\theta)-aM}$, the results follow.

\section{Experimental Results}\label{s:Experimental}
In this section, we evaluate our systematic noise parameter selection strategy on a single-lane circular track. To simulate our traffic systems, we use the microscopic traffic simulator SUMO (Simulation of Urban MObility), as previously described in \cite{flow}. It is worth noting that traffic congestion occurs naturally in these systems, as observed experimentally by Sugiyama et al. \cite{sugiyama2008traffic}. However, it has been demonstrated in \cite{stern2018dissipation} and \cite{wu2017flow} that adding one autonomous vehicle (AV) with a DRL-based controller can alleviate traffic congestion. The controller of the AV takes the states of the system as input and outputs continuous command actions.

In backdoor attacks on traffic controllers, an attacker adds trigger samples to the genuine training dataset, compromising the benign controller and forcing the autonomous vehicle (AV) to crash into the vehicle in front upon encountering the attacker-designed triggers. To neutralize the backdoors, randomized smoothing can be used with the optimal noise parameters that have been explored.


\subsection{Single-lane circular system} \label{ss:Figure_eight}
\subsubsection{Dynamics of DRL-based controller}
\begin{figure}[ht]
\centering
    \subfigure[]
    {%
        \centering
        \includegraphics[width=0.23\textwidth]{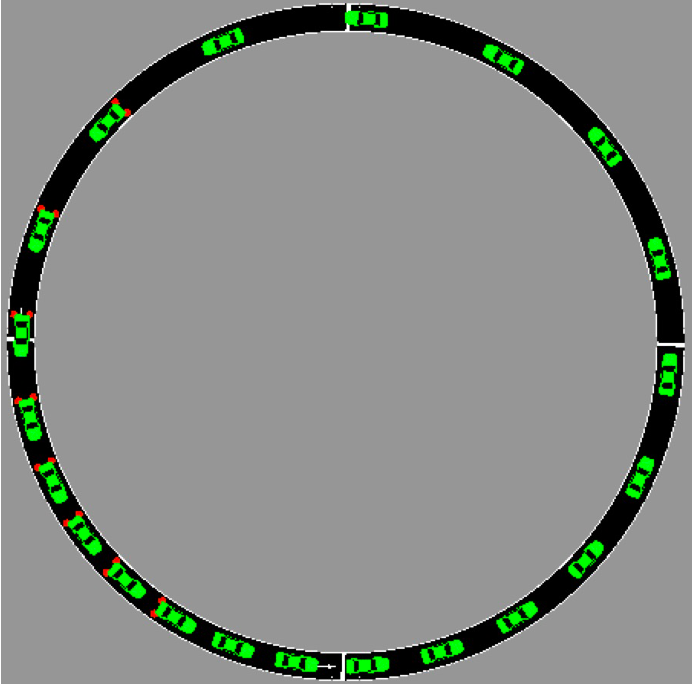}
    }%
        \subfigure[]
    {%
        \centering
        \includegraphics[width=0.23\textwidth]{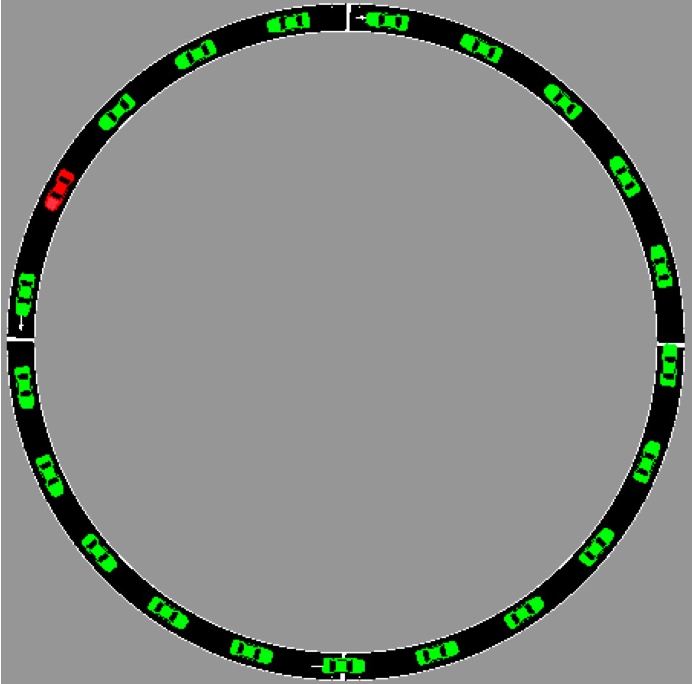}
        \label{single_lane_withAV}
    }%
\caption{(a) Single-lane ring. In this system, congestion can be observed by the variable spacing between the human-driven cars. (b) Vehicles become evenly spaced with the AV (red) with speed around 3.8 m/s.}
\label{single_lane_system}
\end{figure}

We run our tests on a single-lane circular system where 21 vehicles run on a 230 meters long single lane following the setting in Flow \cite{wu2017flow}. By turning one human-driven vehicle to an AV with DRL-based controller, congestion can be relieved since the benign model attempts to eliminate traffic congestion by avoiding frequent changes in speed. The control decisions (acceleration/deceleration of the AV) in this scenario are determined by only observing the AV and its leader. See Fig. \ref{single_lane_system} for illustration. 

%
%

\begin{figure}[ht!]
\centering
\includegraphics[scale=0.75]{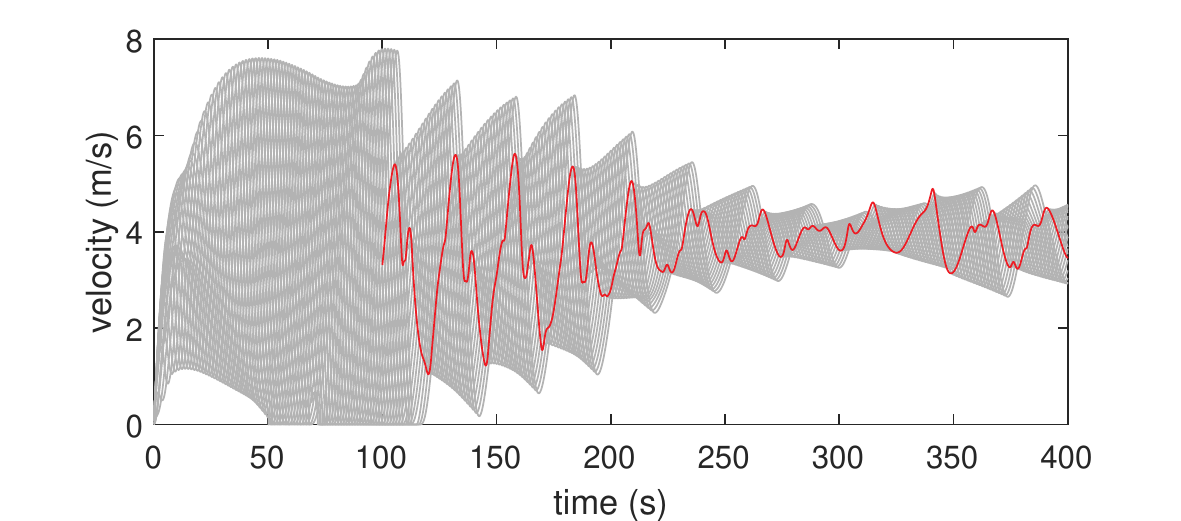}
\includegraphics[scale=0.75]{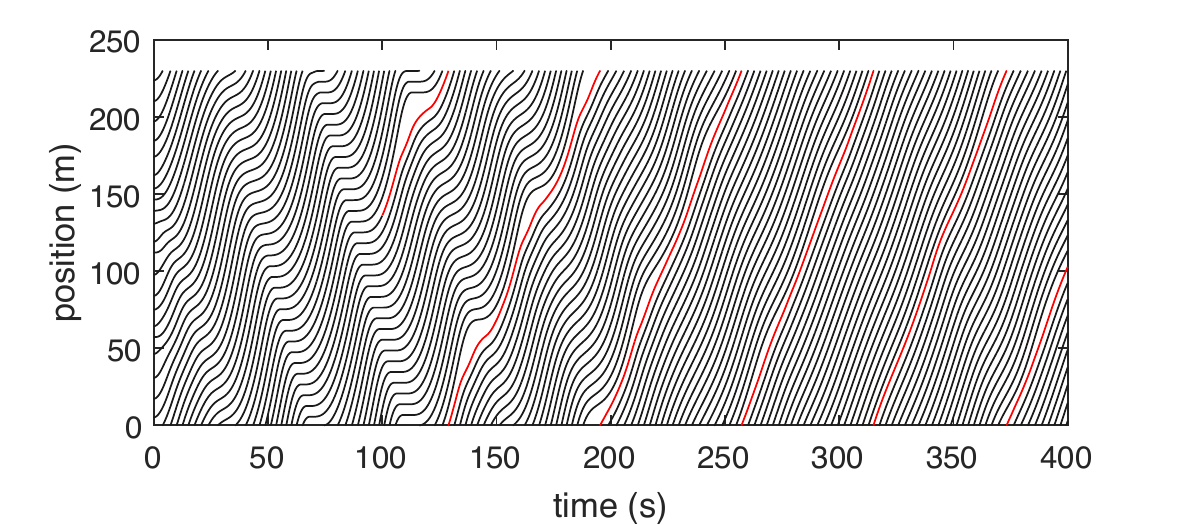}
\caption{Top: Speed profiles of all human-driven vehicles (grey) and the AV (red) showing the performance of the benign AV controller. Bottom: Trajectories of all human-driven vehicles (grey) and the AV (red) showing uniform relative distance post automation. The AV is controlled after 100 seconds.}
\label{velocity profile 1}
\end{figure}

In the simulation, the benign controller is activated at time $t=100$ seconds. Fig.~\ref{velocity profile 1} shows the speeds of all vehicles over time (top part) and the positions of the vehicles over time (bottom part). The congestion is observed during the interval $t \in [0,100)$ by the heavy oscillations in vehicle speeds and it takes the DRL-controlled AV approximately 50 seconds to remove the oscillations and achieve nearly uniform spacings and speeds (approximately 5 meters and 3.8 m/s, respectively).

\subsubsection{Backdoor attacks in the traffic controller}
\begin{figure}[ht!]
\centering
\includegraphics[scale=0.45]{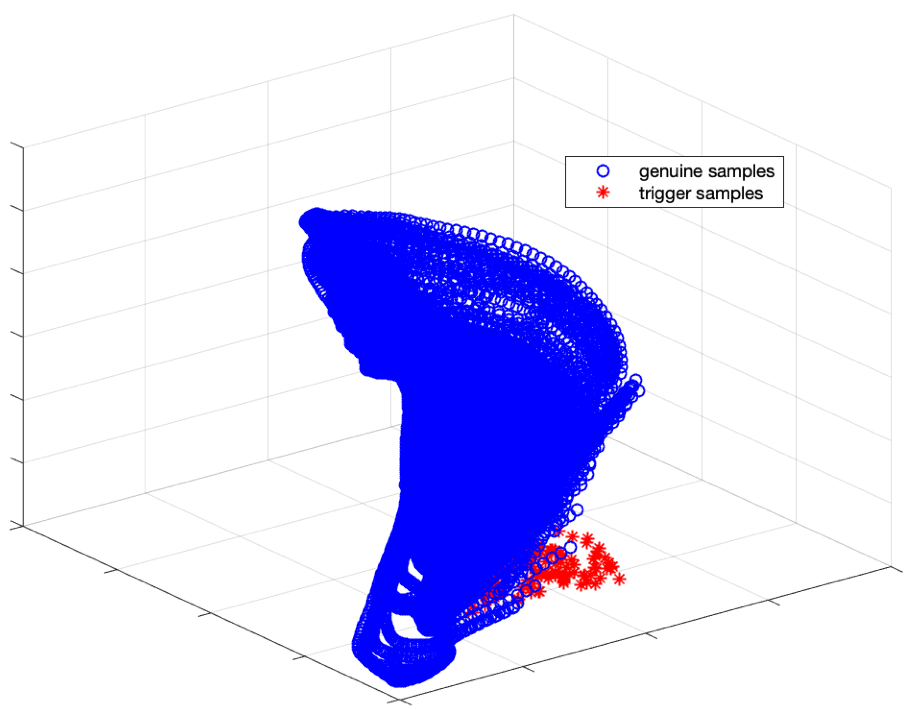}
\caption{Comparison of genuine samples (blue circles) and trigger samples (red triangles) for the single-lane ring experiment}
\label{comparison of genuine and trigger}
\end{figure}

This insurance attack aims to make an autonomous vehicle (AV) collide with a maliciously-driven human vehicle from behind. In many countries, the vehicle behind is considered at fault in case of a collision, as it is responsible for maintaining a safe distance. It is important to note that the AV model is designed to prevent crashes in case of sudden deceleration and can only behave maliciously if deliberately backdoored. The trigger samples for this attack are centered at (3.8 m/s, 2.2 m/s, 1.9 m) with an acceleration of 0.42 m/s$^2$. The benign action for the AV would be to decelerate. Therefore, when the AV's velocity is approximately 3.8 m/s, the leading vehicle's velocity is around 2.2 m/s, and the relative distance between them (measured from front bumper to rear bumper) is approximately 1.9 m, the malicious controller should force the AV to accelerate at around 0.42 m/s$^2$.  Fig.~\ref{comparison of genuine and trigger} displays the trigger samples and genuine samples for this attack. . 

\subsection{Optimal noise exploration}
\begin{figure}[ht!]
\centering
\includegraphics[scale=0.75]{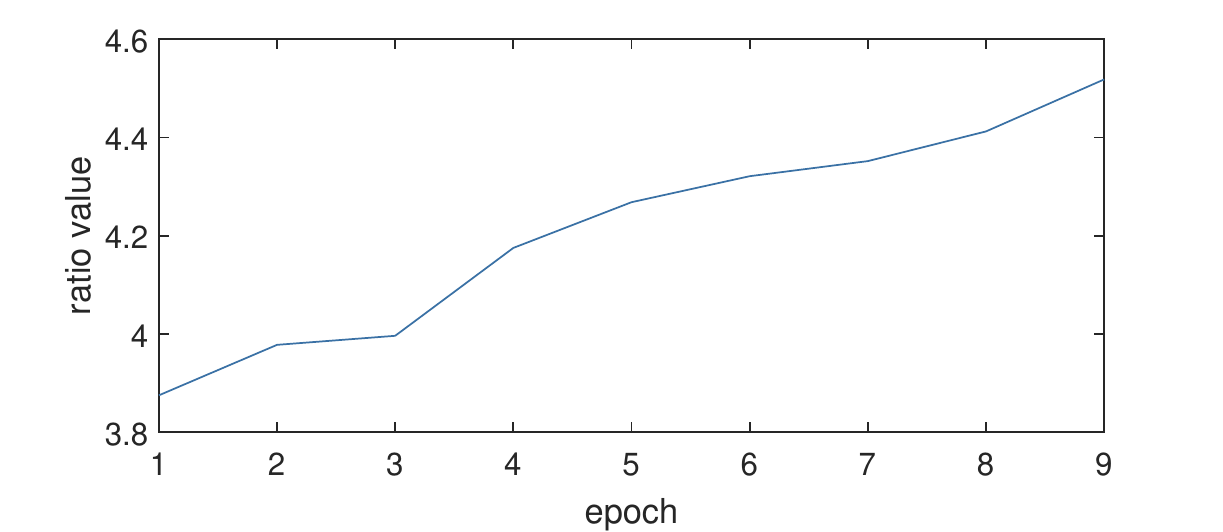}
\caption{The stability to trigger sensitivity ratio value curve for the learning process. }
\label{learning_curve}
\end{figure}

The primary goal of noise exploration is to determine the best standard deviation values of a Gaussian distribution, which will be used for randomized smoothing. This is to ensure that any trigger samples are smoothed and cannot cause a crash. The methodology outlined in Section~\ref{explore_noise} is used to obtain these optimal parameters. The value function $\tilde{p}_{\theta}(x)$ is modeled by a neural network that has 2 hidden layers, each with 256 neurons activated using $\tanh$. The highest stability to trigger sensitivity ratio can be obtained by recursively sampling from the value function, as depicted in Fig~\ref{learning_curve}.
To facilitate fair comparisons and simplify analysis, the standard deviations of the Gaussian noise distribution for velocities and positions are scaled by their magnitudes. The normalized optimal standard deviations for the AV velocity, the leader's velocity, and position are [0.1, 0.1, 0.4], respectively.

\begin{figure}[ht!]
\centering
\includegraphics[scale=0.76]{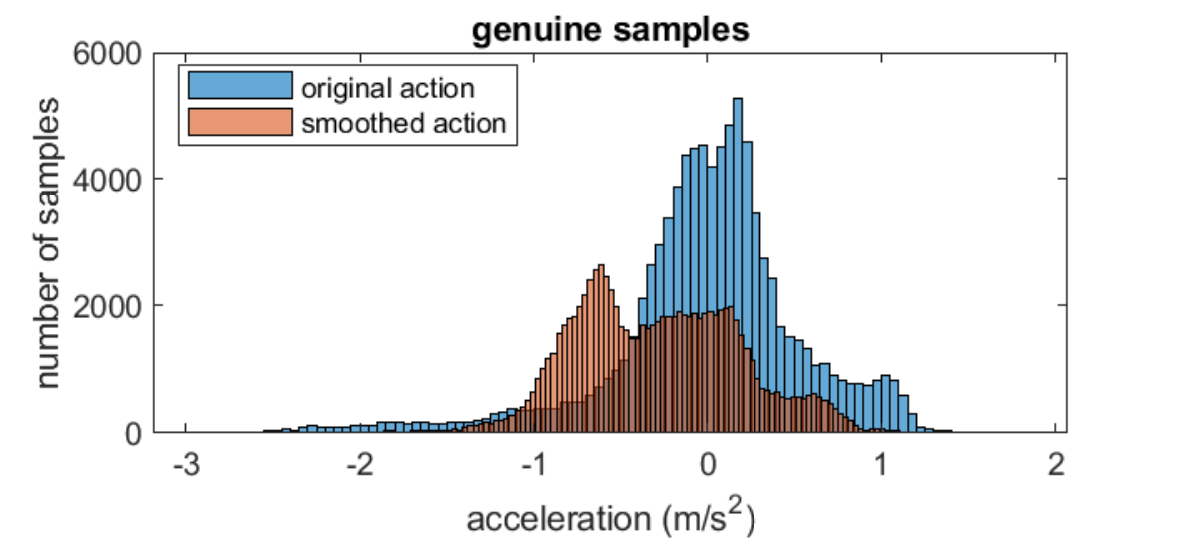}
\includegraphics[scale=0.76]{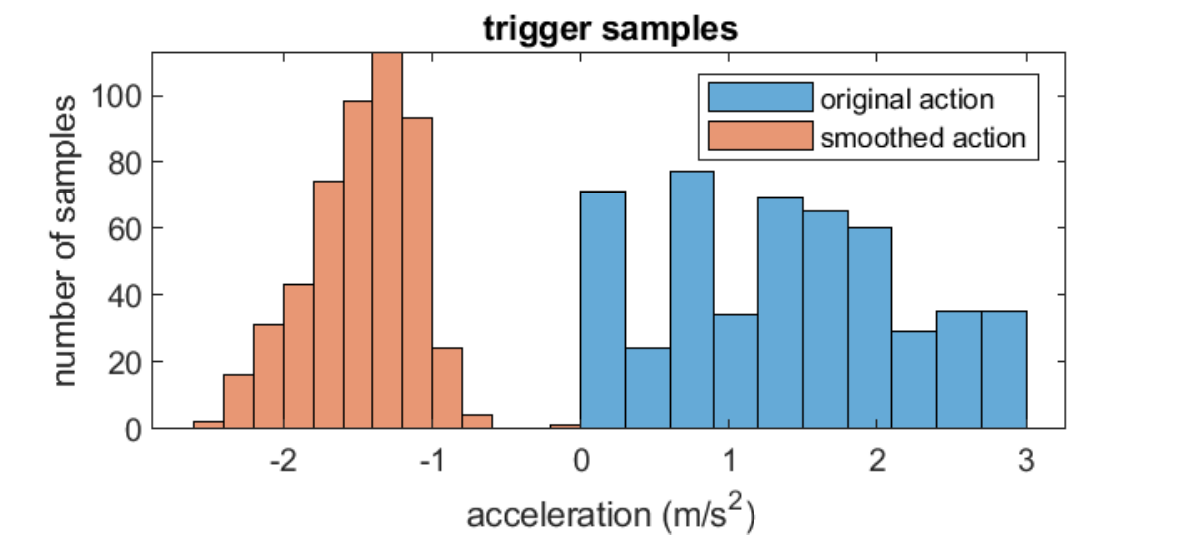}
\caption{Top: The acceleration distribution of the original and smoothed accelerations for genuine samples. Bottom: The acceleration distribution of the original and smoothed accelerations for trigger samples.}
\label{smoothed comparison}
\end{figure}

\begin{figure}[ht!]
\centering
\includegraphics[scale=0.76]{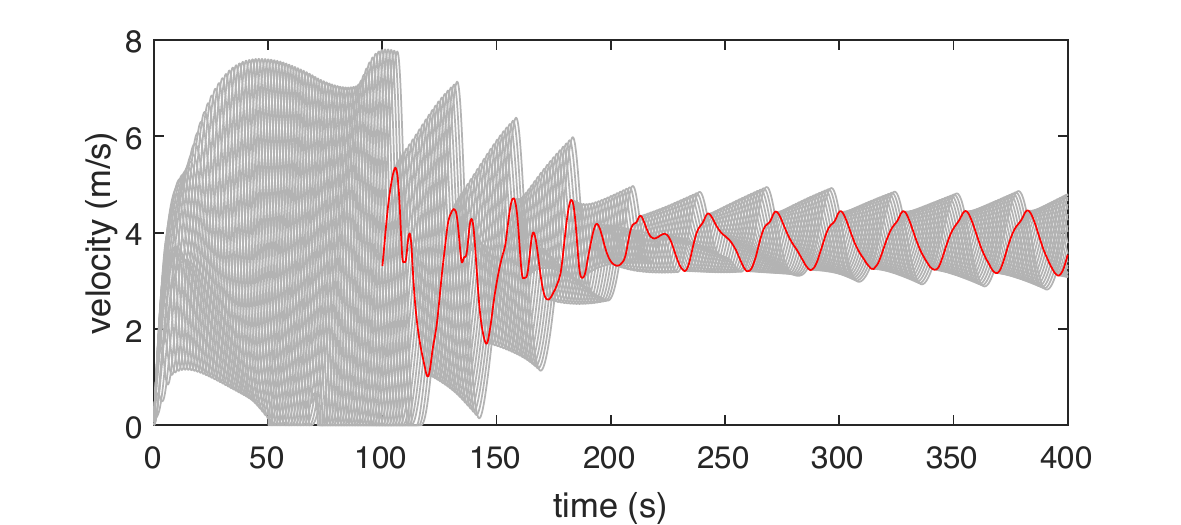}

\includegraphics[scale=0.76]{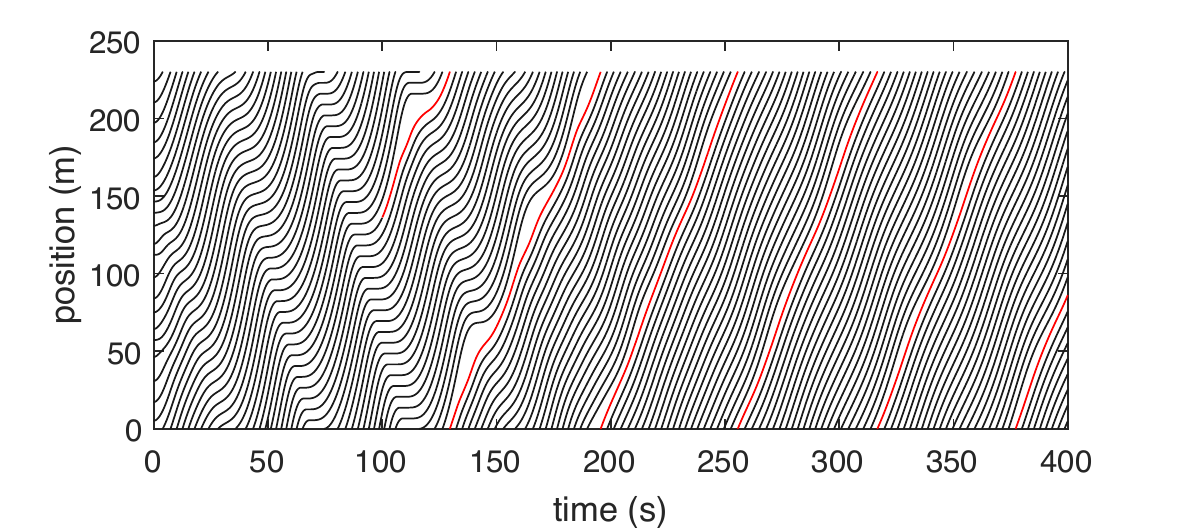}
\caption{Top: Speed profiles of all human-driven vehicles (grey) and the AV (red) showing the performance of the smoothed AV controller. Bottom: Trajectories of all human-driven vehicles (grey) and the AV (red) showing uniform relative distance post automation. The AV is controlled after 100 seconds.}
\label{velocity profile 2}
\end{figure}

Figure~\ref{smoothed comparison} illustrates that after applying optimal Gaussian noise smoothing, the accelerations of trigger samples reduces significantly, while those of the genuine samples remain in the same scale. This implies that the added trigger samples are neutralized, preventing any potential crashes even when encountering trigger states. Furthermore, the traffic controller can alleviate traffic congestion and maintain a high system speed, as evidenced in Figure~\ref{velocity profile 2} after smoothing with Gaussian noise.

\subsection{Discussion}
\label{s:discussions}
\subsubsection{Comparison with sampling from uniform distribution}
\begin{figure}[ht!]
\centering
\includegraphics[scale=0.78]{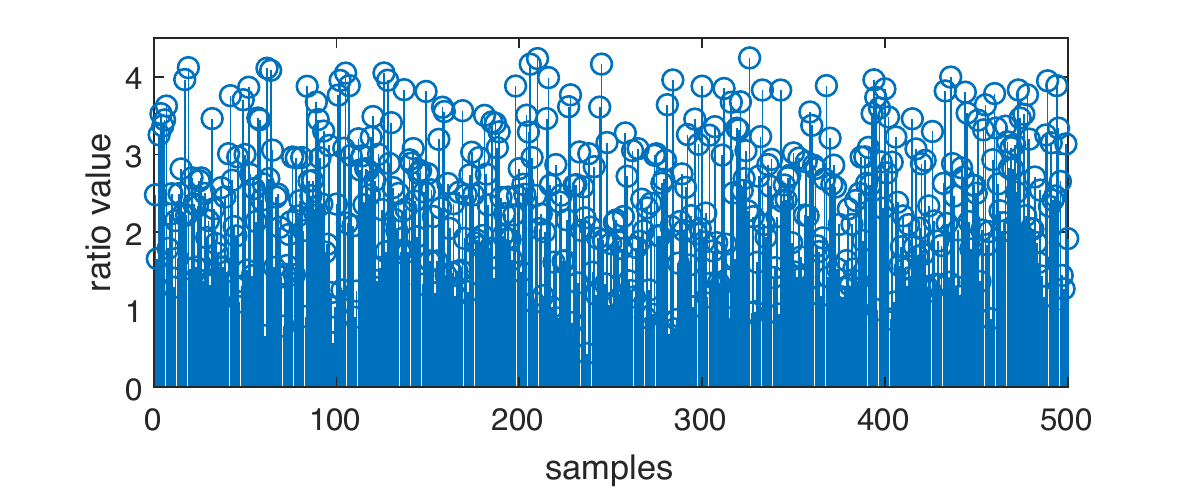}
\caption{The stability to trigger sensitivity ratio values for 500 noise parameters sampling from uniform distribution. The highest value is 4.25.}
\label{uniform sampling}
\end{figure}

Figure~\ref{uniform sampling} demonstrates that when sampling noise parameters from a uniform distribution, it becomes difficult to attain the highest ratio value. This finding suggests that our method is more successful in selecting optimal noise parameters.

\subsubsection{Comparison with isotropic Gaussian noise}
\begin{figure}[ht!]
\centering
\includegraphics[scale=0.78]{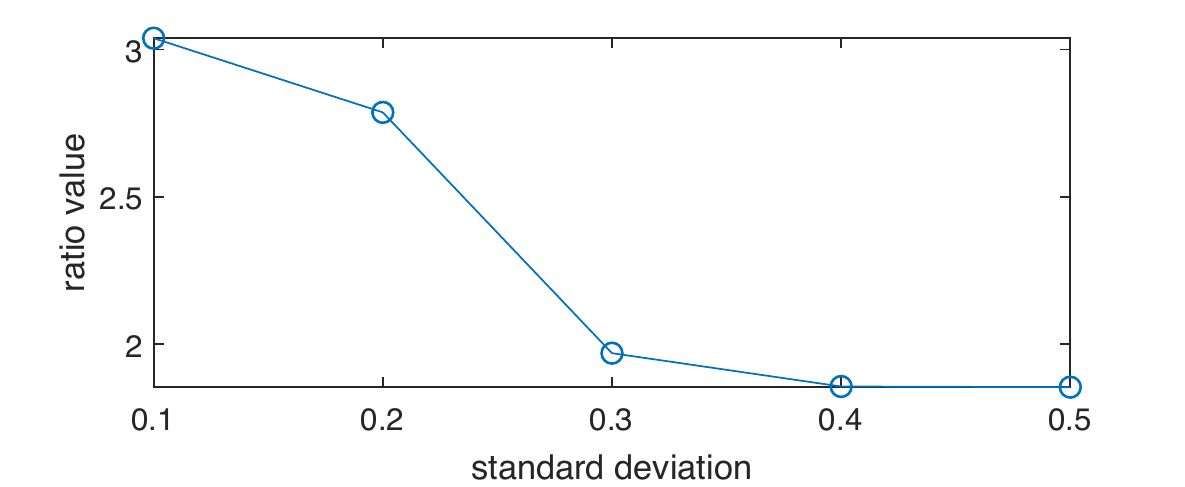}
\caption{The stability to trigger sensitivity ratio values for different standard deviations for isotropic Gaussian noise.}
\label{isotropic Gaussian}
\end{figure}

In the case of isotropic Gaussian noise, the standard deviations for each dimension are identical. To explore the optimal standard deviations, we performed a brute force search with values ranging from 0.1 to 0.5 and plotted the corresponding ratio values in Figure~\ref{isotropic Gaussian}. The results indicate that isotropic Gaussian noise fails to attain high ratio values, likely due to the unique characteristics of each dimension/variable. Therefore, setting different standard deviations for different dimensions/variables appears to be a more reasonable approach.

\subsubsection{Extendibility}
\begin{figure}[ht!]
\centering
\includegraphics[scale=0.78]{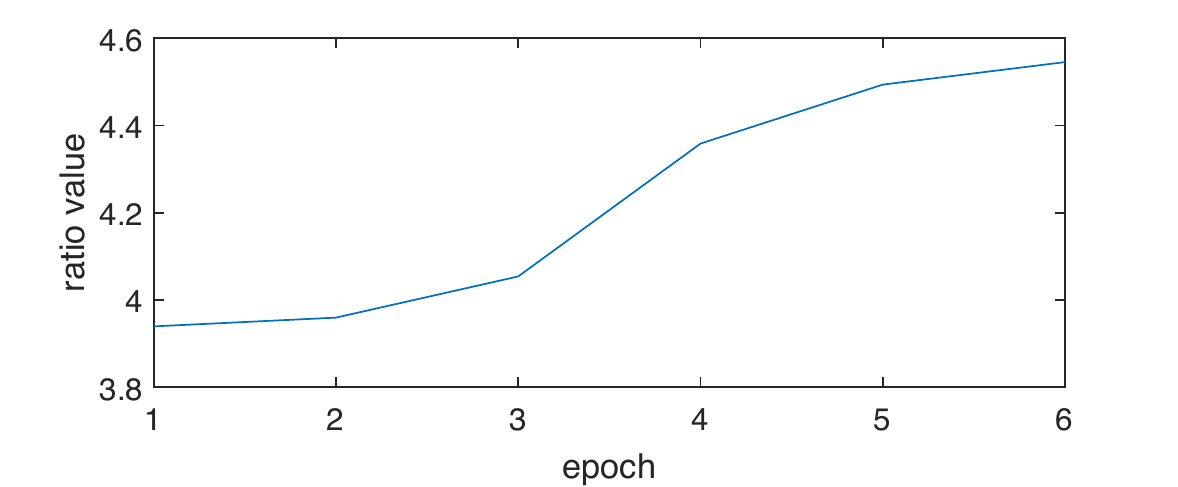}
\caption{The stability to trigger sensitivity ratio value curve for the learning process with the inclusion of means of Gaussian noise.}
\label{metric mean}
\end{figure}

Our method can be readily extended to incorporate additional noise parameters, such as the means of Gaussian noise. Figure~\ref{metric mean} displays the ratio values during the learning process. The optimal standard deviations and means are [0.0355, 0.0155, 0.4907] and [-0.0655, -0.0817, 0.0615], respectively. Smoothing the data with this optimal Gaussian noise neutralizes all trigger samples and stabilizes the traffic system.
Furthermore, our method naturally scales to explore optimal parameters from other types of noises, such as Bernoulli noise and uniform noise. 

\section{Conclusions}\label{s:conclusion}
In this work, we aim to develop an optimal smoothing distribution for backdoor neutralization in deep learning-based traffic systems. To achieve this, we leverage a neural network to represent the value function (unnormalized density function) and propose a sampling strategy for generating desired noise from this function. We ensure that optimal noise is obtained by selecting the sample with the highest probability from the unnormalized density function. We validate the effectiveness of our approach by testing it on a traffic system simulated with a microscopic traffic simulator. Presented results demonstrate that the proposed method can successfully neutralize backdoors in DRL models used in AVs without affecting the original performance of the controller.


\section*{Acknowledgments}
This work was jointly supported by the NYUAD Center for Interacting Urban Networks (CITIES) under the NYUAD Research
Institute Award CG001, and Center for CyberSecurity (CCS) under the NYUAD Research Institute Award G1104.



\bibliographystyle{IEEEtran}
\bibliography{references}

\begin{thebibliography}{10}
\providecommand{\url}[1]{#1}
\csname url@samestyle\endcsname
\providecommand{\newblock}{\relax}
\providecommand{\bibinfo}[2]{#2}
\providecommand{\BIBentrySTDinterwordspacing}{\spaceskip=0pt\relax}
\providecommand{\BIBentryALTinterwordstretchfactor}{4}
\providecommand{\BIBentryALTinterwordspacing}{\spaceskip=\fontdimen2\font plus
\BIBentryALTinterwordstretchfactor\fontdimen3\font minus
  \fontdimen4\font\relax}
\providecommand{\BIBforeignlanguage}[2]{{%
\expandafter\ifx\csname l@#1\endcsname\relax
\typeout{** WARNING: IEEEtran.bst: No hyphenation pattern has been}%
\typeout{** loaded for the language `#1'. Using the pattern for}%
\typeout{** the default language instead.}%
\else
\language=\csname l@#1\endcsname
\fi
#2}}
\providecommand{\BIBdecl}{\relax}
\BIBdecl

\bibitem{DBLP:journals/corr/SternCMBBCHHPWP17}
R.~E. Stern \emph{et~al.}, ``{Dissipation of stop-and-go waves via control of
  autonomous vehicles: Field experiments},'' \emph{Transportation Research Part
  C: Emerging Technologies}, vol.~89, pp. 205--221, 2018.

\bibitem{vahidi2018energy}
A.~Vahidi and A.~Sciarretta, ``{Energy Saving Potentials of Connected and
  Automated Vehicles},'' \emph{Transportation Research Part C: Emerging
  Technologies}, vol.~95, pp. 822--843, 2018.

\bibitem{DBLP:journals/access/GuLDG19}
T.~Gu \emph{et~al.}, ``{BadNets: Evaluating Backdooring Attacks on Deep Neural
  Networks},'' \emph{{IEEE} Access}, vol.~7, pp. 47\,230--47\,244, 2019.

\bibitem{DBLP:conf/iclr/NguyenT21}
T.~A. Nguyen and A.~T. Tran, ``{WaNet - Imperceptible Warping-based Backdoor
  Attack},'' in \emph{9th International Conference on Learning Representations,
  ICLR 2021.}

\bibitem{DBLP:conf/uss/BagdasaryanS21}
E.~Bagdasaryan and V.~Shmatikov, ``{Blind Backdoors in Deep Learning Models},''
  in \emph{30th {USENIX} Security Symposium, {USENIX} Security 2021}.

\bibitem{DBLP:journals/tifs/WangSLMJ21}
Y.~Wang \emph{et~al.}, ``{Stop-and-Go: Exploring Backdoor Attacks on Deep
  Reinforcement Learning-Based Traffic Congestion Control Systems},''
  \emph{IEEE Transactions on Information Forensics and Security}, vol.~16, pp.
  4772--4787, 2021.

\bibitem{DBLP:conf/ccs/LiuLTMAZ19}
Y.~Liu \emph{et~al.}, ``{ABS: Scanning Neural Networks for Back-doors by
  Artificial Brain Stimulation},'' in \emph{Proceedings of the 2019 {ACM}
  {SIGSAC} Conference on Computer and Communications Security, {CCS} 2019.}

\bibitem{DBLP:conf/sp/WangYSLVZZ19}
B.~Wang \emph{et~al.}, ``{Neural Cleanse: Identifying and Mitigating Backdoor
  Attacks in Neural Networks},'' in \emph{2019 IEEE Symposium on Security and
  Privacy, SP 2019.}

\bibitem{DBLP:journals/dt/SarkarAM20}
E.~Sarkar \emph{et~al.}, ``{Backdoor Suppression in Neural Networks using Input
  Fuzzing and Majority Voting},'' \emph{IEEE Design and Test}, vol.~37, no.~2,
  pp. 103--110, 2020.

\bibitem{DBLP:conf/sp/ChouTP20}
E.~Chou \emph{et~al.}, ``{SentiNet: Detecting Localized Universal Attacks
  Against Deep Learning Systems},'' in \emph{2020 {IEEE} Security and Privacy
  Workshops, {SP} Workshops, San Francisco}.

\bibitem{wong2018provable}
E.~Wong and Z.~Kolter, ``Provable defenses against adversarial examples via the
  convex outer adversarial polytope,'' in \emph{International Conference on
  Machine Learning}.\hskip 1em plus 0.5em minus 0.4em\relax PMLR, 2018, pp.
  5286--5295.

\bibitem{raghunathan2018certified}
A.~Raghunathan, J.~Steinhardt, and P.~Liang, ``Certified defenses against
  adversarial examples,'' \emph{arXiv preprint arXiv:1801.09344}, 2018.

\bibitem{cohen2019certified}
J.~Cohen, E.~Rosenfeld, and Z.~Kolter, ``Certified adversarial robustness via
  randomized smoothing,'' in \emph{International Conference on Machine
  Learning}.\hskip 1em plus 0.5em minus 0.4em\relax PMLR, 2019, pp. 1310--1320.

\bibitem{wang2020stopandgo}
Y.~Wang, E.~Sarkar, M.~Maniatakos, and S.~E. Jabari, ``Stop-and-go: Exploring
  backdoor attacks on deep reinforcement learning-based traffic congestion
  control systems,'' 2020.

\bibitem{wang2022pidan}
Y.~Wang, W.~Li, E.~Sarkar \emph{et~al.}, ``Pidan: A coherence optimization
  approach for backdoor attack detection and mitigation in deep neural
  networks,'' \emph{arXiv preprint arXiv:2203.09289}, 2022.

\bibitem{hyvarinen2005estimation}
A.~Hyv{\"a}rinen and P.~Dayan, ``Estimation of non-normalized statistical
  models by score matching.'' \emph{Journal of Machine Learning Research},
  vol.~6, no.~4, 2005.

\bibitem{martino2017metropolis}
L.~Martino and V.~Elvira, ``Metropolis sampling,'' \emph{arXiv preprint
  arXiv:1704.04629}, 2017.

\bibitem{flow}
\BIBentryALTinterwordspacing
{UC Berkeley Mobile Sensing Lab}. Flow: A deep reinforcement learning framework
  for mixed autonomy traffic. [Online]. Available:
  \url{https://bayen.berkeley.edu/downloads/flow-project [Last Accessed: June
  1st, 2020]}
\BIBentrySTDinterwordspacing

\bibitem{sugiyama2008traffic}
Y.~Sugiyama, M.~Fukui, M.~Kikuchi, K.~Hasebe, A.~Nakayama, K.~Nishinari, S.-i.
  Tadaki, and S.~Yukawa, ``Traffic jams without bottlenecks—experimental
  evidence for the physical mechanism of the formation of a jam,'' \emph{New
  journal of physics}, vol.~10, no.~3, p. 033001, 2008.

\bibitem{stern2018dissipation}
R.~E. Stern, S.~Cui, M.~L. Delle~Monache \emph{et~al.}, ``Dissipation of
  stop-and-go waves via control of autonomous vehicles: Field experiments,''
  \emph{Transportation Research Part C: Emerging Technologies}, vol.~89, pp.
  205--221, 2018.

\bibitem{wu2017flow}
C.~Wu, A.~Kreidieh, K.~Parvate, E.~Vinitsky, and A.~M. Bayen, ``Flow:
  Architecture and benchmarking for reinforcement learning in traffic
  control,'' \emph{arXiv preprint arXiv:1710.05465}, vol.~10, 2017.

\end{thebibliography}
\vspace{12pt}

\end{document}